\documentclass[review]{elsarticle}

\usepackage{hyperref}

\journal{Journal of \LaTeX\ Templates}

\begin{document}
\begin{frontmatter}

\title{A Simple and Effective Method of Cross-Lingual Plagiarism Detection}

\author[ispaddress,rauaddress]{Karen Avetisyan\corref{cor}}
\ead{karavet@ispras.ru}
\cortext[cor]{Corresponding author}
\author[ispaddress,rauaddress]{Arthur Malajyan}
\ead{malajyanarthur@ispras.ru}
\author[ispaddress,rauaddress]{Tsolak Ghukasyan}
\ead{tsggukasyan@ispras.ru}
\author[ispaddress,trusted]{Arutyun Avetisyan}
\ead{arut@ispras.ru}

\address[ispaddress]{Institute for System Programming of the Russian Academy of Sciences, Alexander Solzhenitsyn, 25, Moscow, Russia, 109004}
\address[rauaddress]{Russian-Armenian University, Hovsep Emin, 123, Yerevan, Armenia, 0051}
\address[trusted]{ISP RAS Research Center for Trusted Artificial Intelligence, Alexander Solzhenitsyn, 25, Moscow, Russia, 109004}





\begin{abstract}
We present a simple cross-lingual plagiarism detection method applicable to a large number of languages.
The presented approach leverages open multilingual thesauri for candidate retrieval task and pre-trained multilingual BERT-based language models for detailed analysis.
The method does not rely on machine translation and word sense disambiguation when in use, and therefore is suitable for a large number of languages, including under-resourced languages.
The effectiveness of the proposed approach is demonstrated for several existing and new benchmarks, achieving state-of-the-art results for French, Russian, and Armenian languages.
\end{abstract}


\end{frontmatter}

\section{Introduction}\label{sec1}

The automation of plagiarism detection is an essential part of ensuring integrity and fair promotion in academic circles and other spheres of life. 
There have been significant advances in monolingual plagiarism detection, where plagiarism is detected between texts of the same language. 
A lot of research also has been dedicated to solving the problem of cross-lingual plagiarism detection \citep{eiselt2009overview, potthast:2010c,potthast2011overview}.
At the same time, there is notable lack of research and resources for under-resourced languages, as highlighted in \citep{muhammad2020mono}.

Multiple approaches exist for overcoming the language difference for cross-lingual plagiarism detection.
Using machine translation to bring all documents to the same language is the most obvious solution, and it has been successfully employed in many works \citep{kent2010web, sanchez2014winning, muneer2019cleu, bakhteev2019crosslang, kuznetsova2021methods}.
However, the solution relies on a quality machine translation system, which is not available for many language pairs.
Apart from that, translation of entire documents is a time-consuming and expensive process.
There are also corpora-based approaches which leverage parallel texts in different languages to train methods of mapping documents to vectors in a shared space \citep{10.1007/978-3-540-78646-7_51, barron2008cross}. 
These are also not suitable for most language pairs because of the lack of relevant corpora.
Thesauri-based approaches provide alternatives to using machine translation and corpora-based methods, such as using cross-lingual word embeddings, multilingual semantic networks to overcome the language barrier.
This work adopts a similar approach.

Combining the knowledge in multilingual thesauri is a fundamental part of these methods. 
\citep{franco2016systematic} show that using word sense disambiguation is very effective in that regard. 
However, the disambiguation task can be challenging, especially for lexically rich and under-resourced languages. 
Another approach is using fuzzy search, as shown in \cite{ferrero2017usingword}.
They assume that if there is enough textual context, the real concepts will have a considerably higher concept mass and the error will be reduced, which is unfortunately not always the case, especially for lexically rich languages.
Therefore, in this work we propose an alternative approach of merging the thesauri based on simple frequency-based approximation instead of word sense disambiguation.
The proposed plagiarism detection approach displays state-of-the-art results on French, Russian benchmarks and is also demonstrated to be effective on under-resourced languages such as Armenian.

At the detailed analysis step, we compare passages of texts in their original languages, using pre-trained multilingual masked language model to determine their semantic similarity.
The idea came from the success of BERT-based language models in the task of paraphrase detection \citep{devlin2018bert, liu2019roberta}, which is in nature very similar to the target task, and the emergence of multilingual versions of these models \citep{conneau2017word}.

The main contributions of this work are:
\begin{enumerate}
    \item An efficient and accurate method of cross-lingual plagiarism detection that does not use machine translation at runtime and does not rely on word sense disambiguation.

    \item A method of constructing multilingual word clusters from a multilingual thesaurus for the cross-lingual information retrieval of under-resourced languages, also competitive for large languages.


    \item A new test dataset for cross-lingual plagiarism detection, covering 5 language pairs.
    
    \item A challenging test dataset for sentence-level cross-lingual semantic similarity analysis, covering 5 language pairs.

\end{enumerate}

The paper is organized as follows: Section 2 reviews the related work, Section 3 describes the proposed approach, in Section 4 we provide the results of performance evaluation, and discuss it in Section 5.

We publish all presented datasets and models on GitHub.\footnote{\url{https://github.com/1998karen/Cross-Lingual-Plagiarism-Detection}}

\section{Related Work}\label{sec2}

In this section, we present some of the methods previously used for the cross-language plagiarism detection task. 
Similar to monolingual systems, methods in cross-lingual settings also mainly use a  two-staged approach \citep{Potthast:2011}. 
The first stage is the retrieval of a small number of potentially relevant documents from a large reference collection.
It focuses on reducing the search space for the second stage, which focuses on more detailed comparison of the suspicious and retrieved documents aiming to increase precision. 
The documents in the first stage can be full texts, or smaller fragments such as paragraphs, sentences, or fixed-sized windows.
The second one, detailed analysis, pairwise compares the suspicious document to the source ones, retrieved on the previous step. 
Here, the goal is to align the similar text parts of the analysed documents. The method proposed in this paper also relies on two-staged approach.

\emph{Syntax/Lexical based methods} calculate the similarity without using any multi-lingual data or translation mechanisms, and are mainly used for languages that are syntactically and lexically close to each other.
For example, CL-CNG \citep{McNamee2004CharacterNT} is one such method that uses character n-gram vectors to represent the texts. 

\emph{Thesauri-based methods} are based on transforming texts into language-independent forms, with their following comparisons. 
Thesauri such as BabelNet \citep{NAVIGLI2012217} or EuroWordNet \citep{vossen1998introduction} are multilingual semantic networks of different text units.
For each word, MLPlag \citep{Ceska2008MultilingualPD} extracts language-independent word senses, using EuroWordNet thesaurus, and later directly compares them.
Cross-Language Conceptual Thesaurus based Similarity (CL-CTS) method \citep{10.1007/978-3-642-33247-0_8} transforms documents into vectors using Eurovoc\footnote{\url{https://op.europa.eu/s/vFSH}} conceptual thesaurus, and then uses cosine similarity for comparison of the vectors. 
Another method is a knowledge graph-based (CL-KGA) approach \citep{10.1007/978-3-642-36973-5_66, FRANCOSALVADOR2016550} which uses BabelNet.
These graphs are constructed for each text fragment, using the words' concepts and the paths between them. 
Finally, the similarity is computed with some graph similarity algorithm.
The described methods often rely on word sense disambiguation, which prevents them from being used for under-resourced languages that do not have such tools.

\emph{Corpora based methods} use different information extracted from multi-lingual databases.
The cross-language explicit semantic analysis (CL-ESA) model \citep{10.1007/978-3-540-78646-7_51} is a multi-lingual version of Explicit Semantic Analysis \citep{Gabrilovich:007}.
Assuming that D is a document collection, for each document \textit{d} of that collection, the ESA algorithm associates a concept vector \textbf{d}, which in its turn is constructed using TF-IDF similarity between the considered document and each of the  Wikipedia article concepts.
The CL-ESA method works exactly the same way but instead of using monolingual collection D, it uses a concept-aligned multi-lingual set of collections.
Another method based on parallel corpora is the Cross-Language Alignment-based similarity Analysis (CL-ASA) \citep{barron2008cross}, which utilizes a statistical bilingual dictionary obtained via IBM alignment model 1 \cite{brown1993mathematics}.
The drawback of these methods is the need for training sets, which are available only for a handful of language pairs.

\emph{Machine translation based methods} translate the documents into one language and then use mono-lingual search algorithms. 
Translation plus mono-lingual (T+MA) based approaches were used by many authors \citep{kent2010web, sanchez2014winning, muneer2019cleu, bakhteev2019crosslang, kuznetsova2021methods}.
Similar to corpora-based approaches, using machine translation is not an option for most language pairs, because developing or getting access to an accurate translator would be extremely expensive.

\emph{Word embedding based algorithms.} The methods use the vector representations of the words to compare the texts.
In \citep{ferrero2017usingword} two methods were proposed, CL-CTS-WE uses each word’s top 10 closest word embeddings to build a bag-of-words of that word.
The second one, CL-WES, represents each considered textual unit as a vector sum of the words it consists of. 
Bilingualism of the vectors for both methods is provided by Multivec \citep{berard2016multivec}. 
Another method \citep{roostaee2020cross} based on word embeddings uses bilingual word vectors from \citep{conneau2017word}.

Besides the separate approaches mentioned above, some others like \citep{roostaee2020effective} or \citep{ferrero2017usingword} try to use fusions of those to achieve better results.

The proposed method belongs to thesauri-based methods, but in contrast to the most does not rely on word sense disambiguation. To the best of our knowledge the proposed method is the first one to use Universal WordNet for the multilingual plagiarism detection task.

\section{Proposed Method}\label{sec3}

This section describes the proposed two-step approach for cross-lingual plagiarism detection.
The following subsection describes the process of retrieving a small set of relevant text passages from a large collection of reference documents.
After that we describe the detailed analysis step where the retrieved passages are compared with the suspicious passage using a more precise method. 

Further in text, the terms suspicious and reference documents/passages will be used. Suspicious documents/fragments denote the ones that potentially contain plagiarism. Reference documents/passages denote the ones from which plagiarism can potentially be produced. 

\begin{figure*}[h]
\centering
\includegraphics[width=\textwidth]{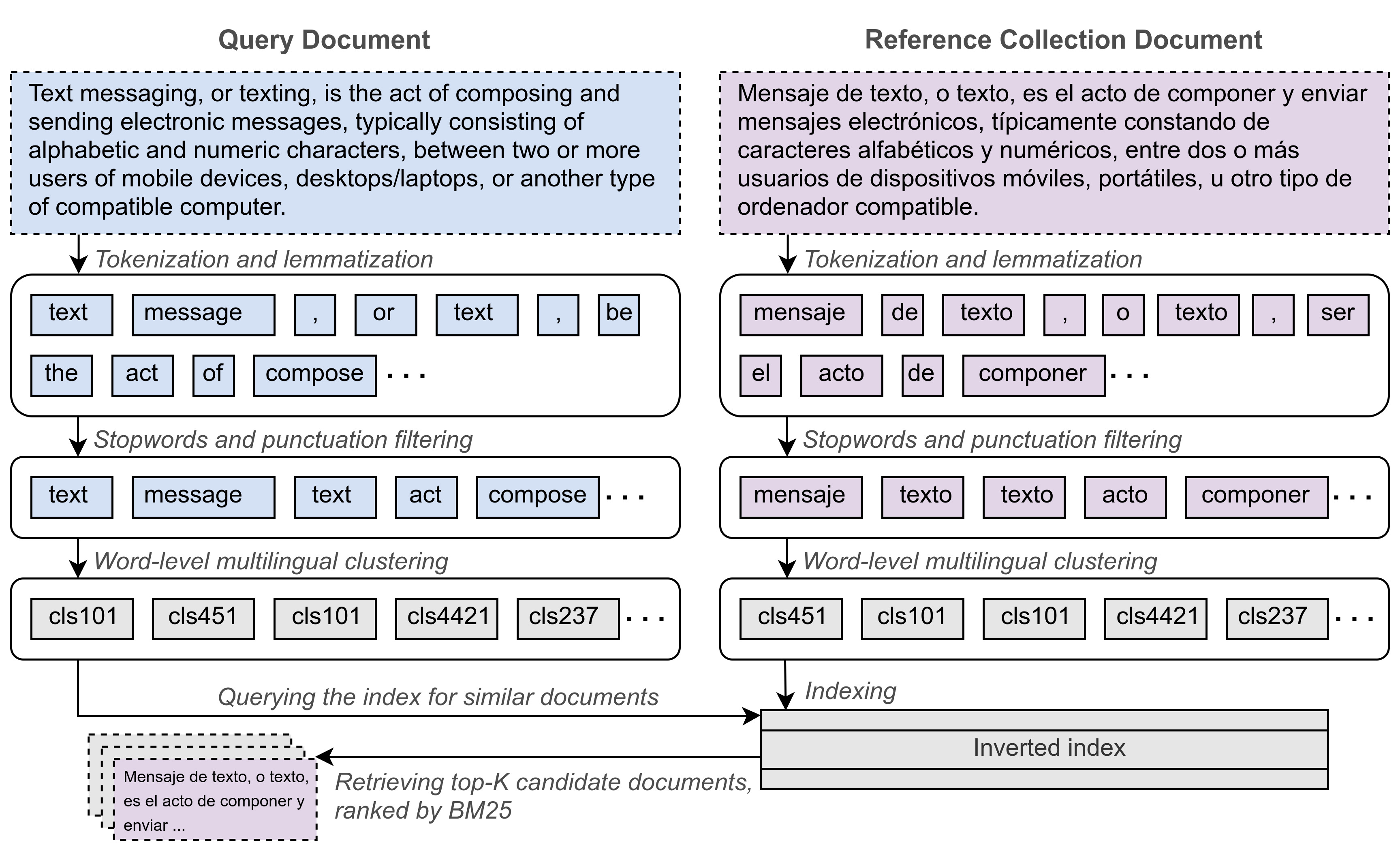}
\caption{The scheme of query and reference collection documents preprocessing, word-level multilingual clustering and relevant fragment searching.}
\label{fig:candidate-retrieval}
\end{figure*}

\subsection{Candidate Retrieval}

The candidate retrieval process is a relevant document extraction process. 
It reduces the number of documents that will go through the computationally expensive detailed analysis step. 
In the proposed method, instead of full texts both suspicious and reference documents are divided into smaller passages of text (for example, paragraphs), and relevant candidate retrieval is done at that level.

We define the task as follows: given a set of text passages $D$ in language $L_1$, a suspicious passage of text $s$ in language $L_2 (L_2 \neq L_1)$, and a relevance ranking method $M$, the goal is to retrieve from set $D$ the most relevant $K$ passages of text for $s$, where $K \ll |D|$.

The method employs an inverted index for implementing fast search. 
Reference documents are divided into smaller text fragments, which are then pre-processed and indexed. 
Using the predefined multilingual thesaurus that maps each word to a set of unique concepts, each paragraph’s word is replaced with the corresponding concept. 
Words that are not attached to any concept remain unchanged. 
Afterward, an inverted index is built for each of those modified paragraphs. 
Suspicious documents are divided into sentences, and the replacement process is applied on the sentence level. This process is described in Section \ref{mwc}.

The search is performed for suspicious replaced sentences trying to find the most similar source modified paragraphs. 
In addition to the multilingual thesaurus, to enhance the search quality, we pre-processed both source and suspicious documents. 
The text was lemmatized and after that all stopwords, numbers, punctuation characters as well as other tokens containing punctuation inside them were removed. 

Described document indexing and search processes were implemented using the open-source enterprise-search platform Apache Solr\footnote{\url{https://solr.apache.org/}}.
Okapi BM25 was utilized as the ranking function \cite{robertson1995okapi}.
For tokenization and lemmatization, we used Stanza v1.2.1 \citep{qi2020stanza}.

The candidate retrieval process is shown in \figurename{ \ref{fig:candidate-retrieval}}.

\subsubsection{Multilingual Word Clusters} \label{mwc}

\begin{figure*}
\centering
\includegraphics[width=\textwidth]{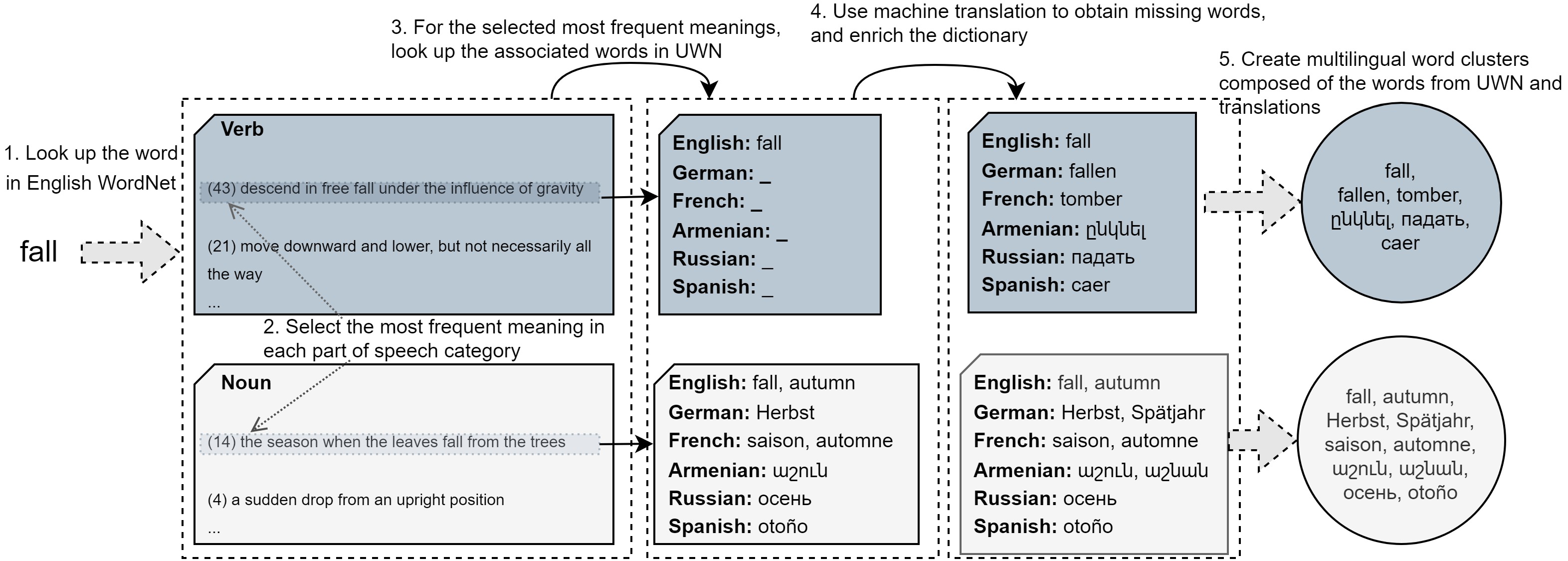}
\caption{The process of preliminary multilingual word cluster creation using Universal WordNet and its addition with the use of machine translation.}
\label{fig:syn-dict}
\end{figure*}

To obtain the correct concept for each word, previously published work relied on word sense disambiguation.
In this work we propose circumventing that step and using a word frequency-based approximation to merge thesauri of different languages into one.

Many multilingual thesauri exist that can be used in the abovementioned method. BabelNet \cite{NAVIGLI2012217}, DBNary \cite{serasset2015dbnary}, and various multilingual WordNets provide open access to their data, which could be adapted to create multilingual word clusters for the cross-lingual retrieval task.
Based on the vocabulary size, its coverage of a large number of languages and precision, we decided to use Universal WordNet (UWN) \cite{deMeloWeikum2009UWN} as basis for our experiments.

Within a language, each word is linked to one or more synsets, each corresponding to a unique concept.
When merging UWN's interconnected synsets from different languages, the number of words in created multilingual clusters of concepts grows even more.
The usage of all available synsets for a word leads to a situation when a word appears in many clusters, which negatively affects the precision of retrieval method. 
Therefore, it is important to reduce the number of each word's clusters to the most relevant ones.

In this work, we experiment with several approaches to alleviating that problem.
First, we used the concepts in Princeton WordNet 3.1\footnote{\url{https://wordnet.princeton.edu/}} to create the initial set of word clusters. 
The concepts were extracted for every English word.
Then, we used UWN to populate the clusters with corresponding words in other languages.

Let $L_{UWN}$ be the set of all languages supported by UWN, $C_{UWN}$ be the set of all concepts, universal among all languages, and $w$ some word in English.
With $S^l_{C_w}$ we denote the set of synsets corresponding to all of the concepts of the word $w$ ($C_w \in C_{UWN}$) in language $l \in L$.
Then, we can define the different multilingual cluster creation approaches as follows:





\textbf{All}. For each English word $w$ we merge the synsets $S^l_{C_w}$ for all $l \in L_{UWN}$. 
The synsets that correspond to different parts of speech of the $w$ are merged separately. 
The merged synsets serve as multilingual word clusters. 

\textbf{Top1}. In this approach, we additionally filter the set of concepts based on the frequencies provided in Princeton WordNet.
For each English word $w$, we take only the most frequent concept $C^{Top1}_w$ and merge its corresponding synsets $S^l_{C^{Top1}_w}$ for all $l \in L_{UWN}$.
The most frequent concepts are taken separately for each part-of-speech category of the word $w$.

We call the created multilingual word-to-cluster mapper “Cross-Language synonyms dictionary” (CL-SynDi). 
Since Universal WordNet consists of over 1,500,000 words in over 200 languages, CL-SynDi mapper can be used for building cross-lingual information retrieval systems for all those languages.

\begin{figure}[h]
\centering
\includegraphics[width=0.6\columnwidth]{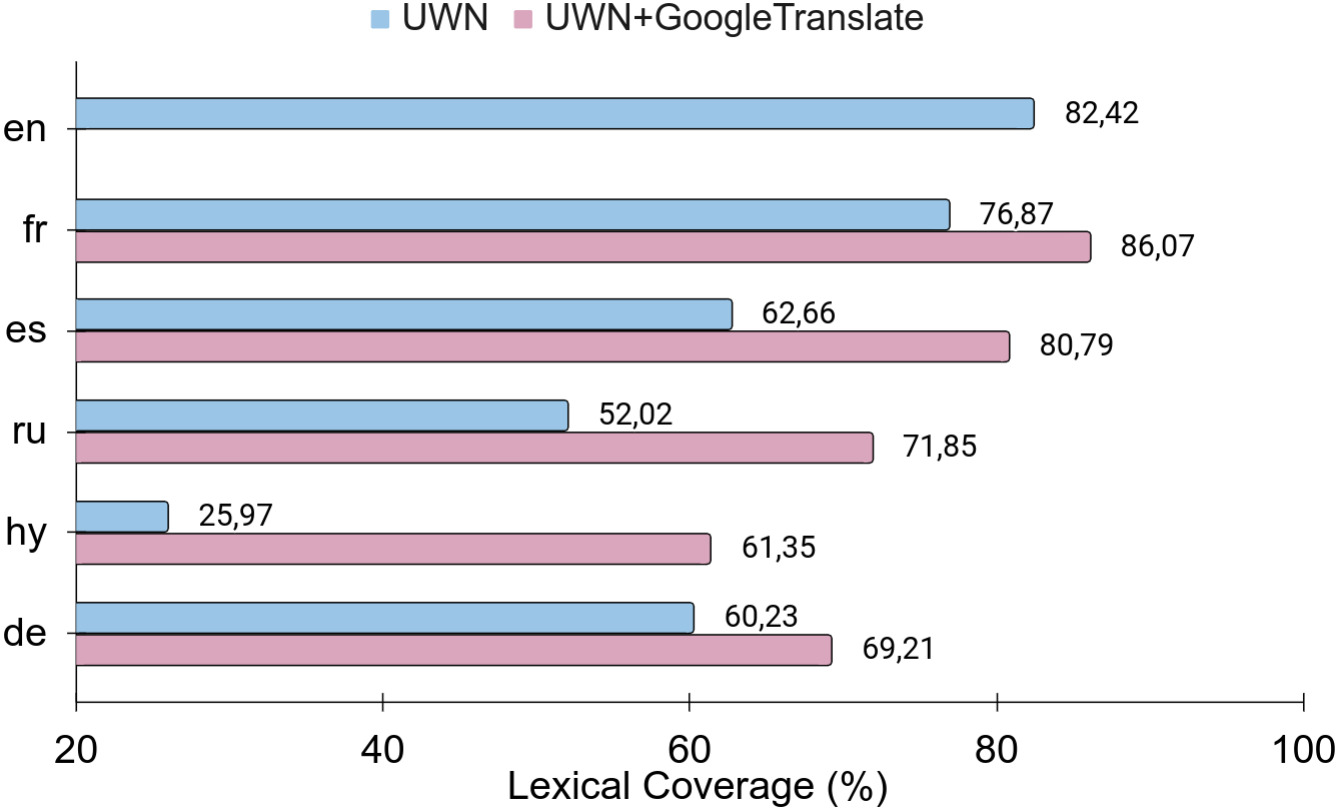}
\caption{The comparison of lexical coverage of the multilingual clusters created with and without the usage of machine translation, computed on 120,000 texts of Wikipedia.}
\label{fig:syn-dict-coverage}
\end{figure}

The analysis of the word-to-concept mapping obtained from UWN revealed many concepts missing the corresponding words for some languages. 
Thus, we manually supplemented the missing meanings (in the way of translation) to the clusters in the respective languages.
Additionally, we enriched the clusters with the words in the languages that were already present.
The translation was processed using Google Chrome translation.
Only the English words were used as a source of translation.
To get the proper translation for the words that can appear as different parts of speech, the words were translated in connection to their parts of speech.
The process of CL-SynDi creation is described in \figurename{ \ref{fig:syn-dict}}.
The final statistics of the obtained dictionaries are shown in \figurename{ \ref{fig:syn-dict-coverage}}. 

While less fine-grained, the proposed approach achieves good results on various cross-lingual benchmarks and as shown for the Armenian language, can be quite effective for under-resourced languages that might not have word sense disambiguation tools.

The created multilingual clusters are published on GitHub\footnote{\url{https://github.com/1998karen/Cross-Lingual-Plagiarism-Detection}}.

\subsection{Detailed Analysis}

After retrieving the top-K relevant fragments, we perform more detailed analysis of their similarity to the suspicious fragment. The goal of this step is to increase the overall precision of the plagiarism detection. 

\begin{figure}[h]
\centering
\includegraphics[width=0.6\columnwidth]{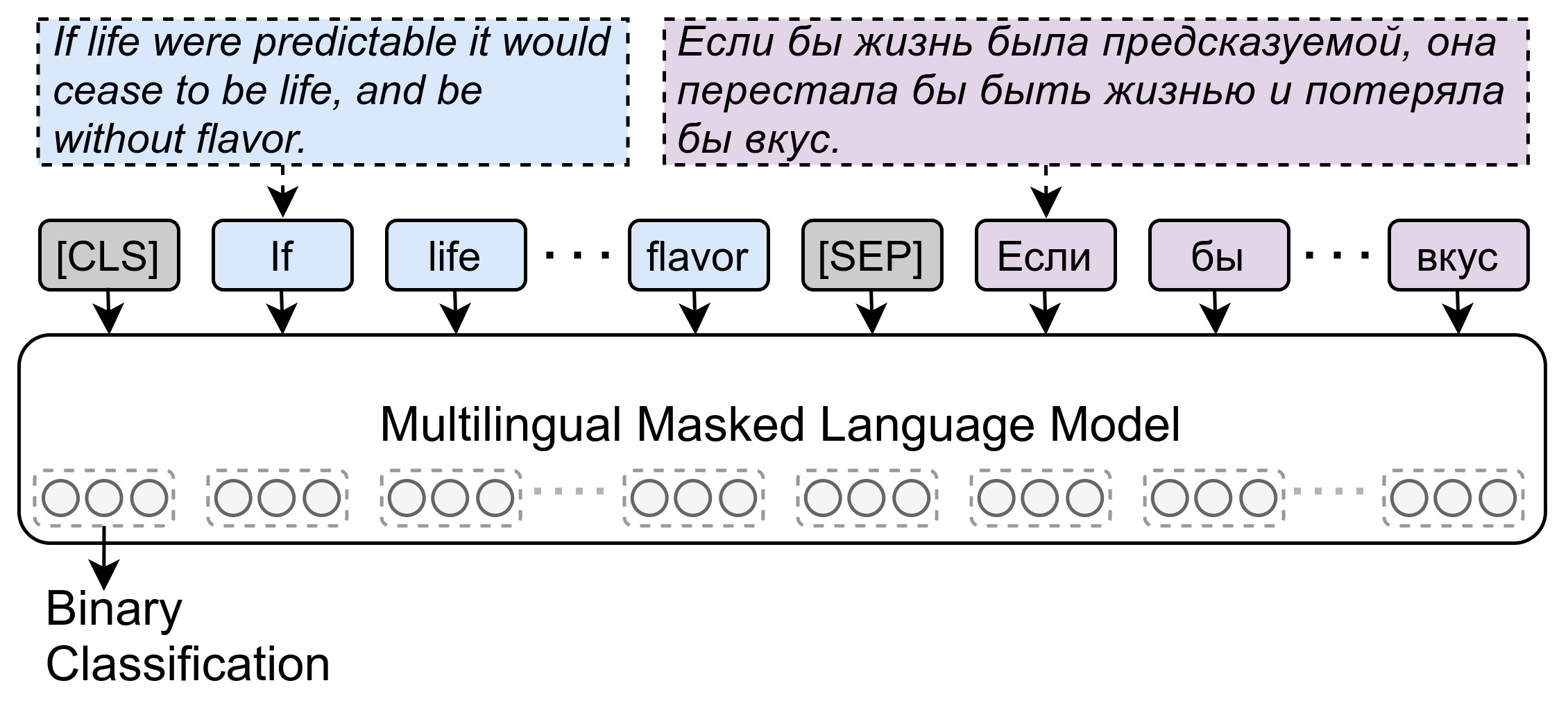}
\caption{The process of classification whether the two text fragments are translation of each other, used at detailed analysis step.}
\label{fig:text-alignment-model}
\end{figure}

In the detailed analysis step, we employ the same method as \citep{zubarev2019cross}. 
The task is defined as a binary classification of whether the two input passages of texts in different languages are a translation of each other.
Similar to \citep{zubarev2019cross}, we also fine-tune a pre-trained multilingual  masked language model and use a pair of sentences as input. 
If multiple candidates are classified as translation, the one with the highest score is selected.
The model and how it works is shown in Figure \ref{fig:text-alignment-model}.
If retrieved and suspicious texts fragments contain more than one sentences, they are divided into sentences, which are then pairwise compared using the fine-tuned language model.

It is worth noting that the defined translation detection task is similar to paraphrase detection.
Taking that into account, we decided to use multilingual BERT-based language models for our task, since they have shown decent results for paraphrase detection. 

To fine tune the language model we used pairs of parallel sentences as positive examples, and as negative examples we used sentences that were similar but not the translation of each other.

\section{Experiments}\label{sec4}

To demonstrate the effectiveness of the proposed approach, we carried out several experiments, evaluating candidate retrieval and detailed analysis steps separately, and also the overall performance on established benchmarks and new test datasets. 
We performed the evaluations for English-Russian, English-French, English-Spanish, English-German, and English-Armenian language pairs.

\subsection{Candidate Retrieval}

To select the best of proposed modifications and synonym dictionaries for candidate retrieval part of the algorithm, we ran several tests.  
We added 10,000 English documents, extracted from Wikipedia, to the index as reference collection. To check the candidate retrieval performance, we then performed search for 316 automatically generated suspicious documents in Russian. All of the suspicious documents contained plagiarised fragments. 

As an evaluation metric of candidate retrieval step \textit{Recall@K} was chosen, which computes the ratio of the relevant documents in the retrieved $K$ documents. 

According to the estimations shown in Table \ref{table:top1}, the best results were achieved by using the word-to-concept mapper that was created using only the meanings with the highest frequency scores.

\begin{table}[h]
\centering
\resizebox{0.7\columnwidth}{!}
{\begin{tabular}{lcccc}
\hline
             & \multicolumn{4}{c}{\textbf{Recall@K}} \\
\textbf{Merge method} & \textbf{K=1} & \textbf{K=5} & \textbf{K=10}  & \textbf{K=50}\\
\hline
All   & 0,624 & 0,728 & 0,762 & 0,832 \\
Top1    & \textbf{0,747} & \textbf{0,827} & \textbf{0,853} & \textbf{0,899} \\
\hline
\end{tabular}}
\caption{Extrinsic evaluation of cross-lingual thesauri merge methods in the candidate retrieval task.}
\label{table:top1}
\end{table}

Additionally, we checked the effectiveness of extending the thesaurus with machine translation (Table \ref{table:UWN-GT}).

\begin{table}[h]
\centering
\resizebox{0.7\columnwidth}{!}
{\begin{tabular}{lcccc}
\hline
                       & \multicolumn{4}{c}{\textbf{Recall@K}} \\
\textbf{Thesaurus}     & \textbf{K=1}   & \textbf{K=5}   & \textbf{K=10}  & \textbf{K=50}\\
\hline
UWN                    & 0,175          & 0,265          & 0,307          & 0,426\\
UWN+GoogleTranslate & \textbf{0,747} & \textbf{0,827} & \textbf{0,853} & \textbf{0,899} \\
\hline
\end{tabular}}
\caption{Extrinsic evaluation of the effectiveness of machine-translated multilingual thesaurus in the retrieval task.}
\label{table:UWN-GT}
\end{table}



\subsubsection{Hyperparameters}

For candidate retrieval, we selected \textit{$K = 50$} to guarantee high level of recall.
For example, \cite{pataki2012new} work for English-Hungarian plagiarism detection also considered up to 50 documents for the retrieval stage.
For Okapi BM25 ranking of the candidates, we used $k1 = 1.2$ and $b = 0.75$.
Suspicious documents were divided into sentences, while reference documents were divided into paragraphs.
This configuration was used in all following experiments. 

\subsection{Detailed Analysis}

XLM-RoBERTa (XLM-R) \citep{conneau2020unsupervised} was chosen as the model to resolve the detailed analysis task. 
XLM-RoBERTa is a multilingual model trained on 100 different languages. 

We created a training dataset to fine-tune the XLM-R model, and then evaluated the performance on several datasets.

\textbf{Training.} To fine-tune XLM-R we automatically generated a dataset, using English Wikipedia and scientific papers on various topics, randomly sampled from Google Scholar. 
The positive examples were generated by automatically translating random sentences from scientific papers and Wikipedia sentences, 4,500 each. 
In total, the training set included 9,000 positive examples, 150 for each language pair. 

As for the negative examples, for each language pair, we took random sentence pairs from English Wikipedia and scientific papers 75 each.
These sentences were then translated into corresponding languages to form sentence pairs for each language pair.
We assume that during the plagiarism detection process, the vast majority of sentence pairs that will rich the detailed analysis step will not have or will have low similarity between their sentences. 
During the data generation process, we checked that no sentence appeared twice in the training set.
Finally, the training set consisted of 18,000 examples.

\textbf{Evaluation}. The trained model was evaluated on existing Negative-1, Negative-4 test datasets \citep{zubarev2019cross} and a new dataset based on MRPC \citep{dolan2005automatically}.
The evaluation results are provided in \textbf{Table \ref{table:da}}.

\textbf{Negative-1, Negative-4.} These test datasets from \citep{zubarev2019cross} consist of English-Russian sentence pairs. 
They were used to fine-tune the model for that language pair only. 
16,000 sentences from Yandex parallel corpus and 4,000 manually translated sentences were used,  as positive examples. Negative examples were obtained using the same 20,000 sentences. Each Russian sentence was compared to each English sentence (except for its own pair) using different similarity measures. For the Negative-1 set, only one negative example was taken for each Russian sentence from the most similar sentences. As for the Negative-4 set, 4 negative examples were taken (one most similar example for each similarity measure).

\textbf{$\nu$MRPC}. The main drawback of existing detailed analysis datasets is that they generate non-translation pairs mostly based on lexical distance.
That approach leaves out certain difficult cases where the sentences are lexically distant but semantically very close.
Based on the similarity of paraphrase and translation detection tasks, came the idea to transform a paraphrase detection dataset for our needs. 
Translation of near-paraphrases from a paraphrase detection dataset would allow to address the lack of challenging negative examples.
For more robust evaluation of the trained detailed analysis model, we created a new test dataset for 5 pairs of languages based on machine translation of the Microsoft Research Paraphrase Corpus (MRPC).
We denote the new dataset $\nu$MRPC. 

\begin{figure}[h]
\centering
\includegraphics[width=0.6\columnwidth]{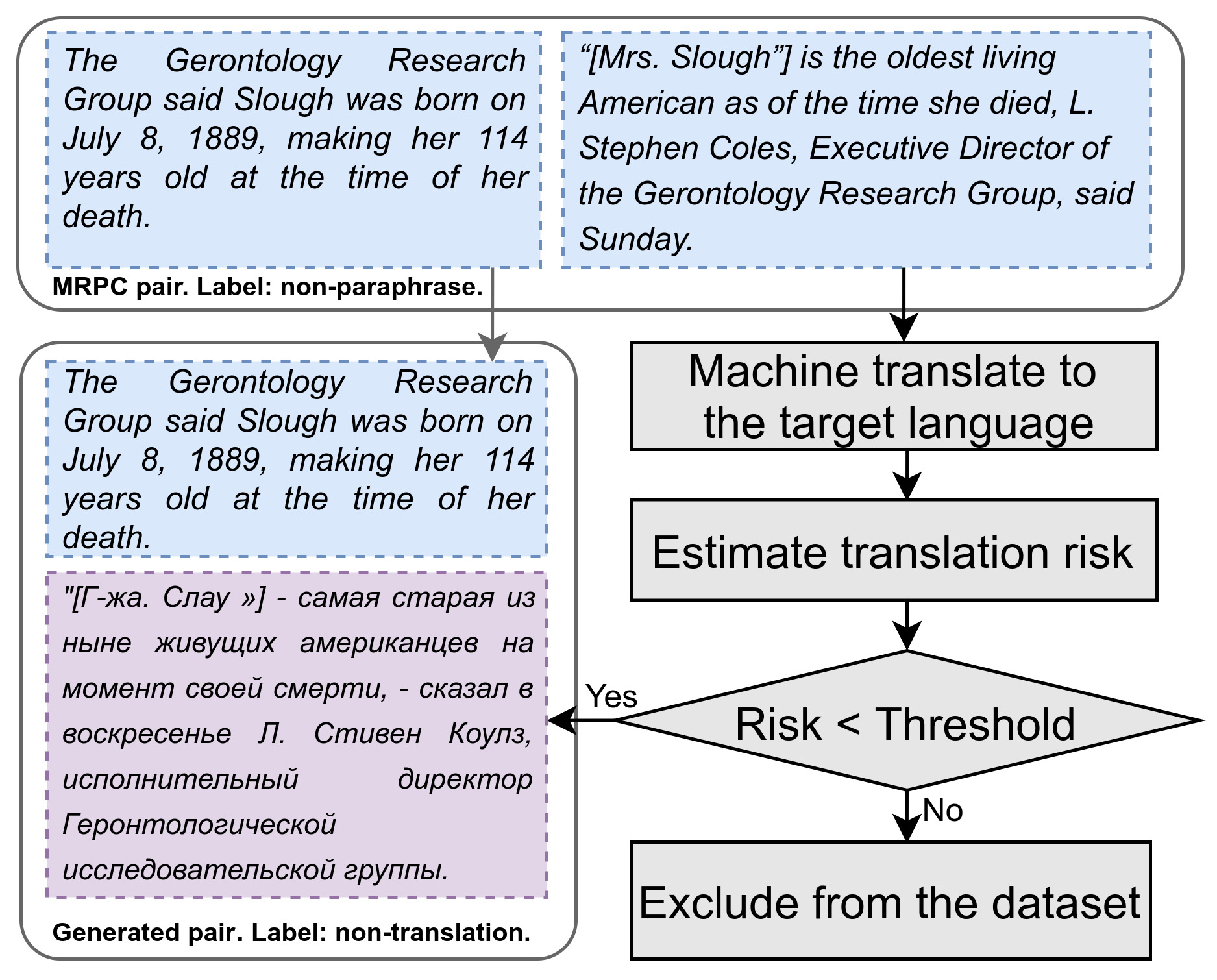}
\caption{Detailed analysis challenging test data generation translating one of the sentence in MRPC paraphrase pairs.}
\label{fig:TA-data-generation}
\end{figure}

The process of creating $\nu$MRPC is described in Figure \ref{fig:TA-data-generation}. 
For each pair from MRPC dataset, one of the sentences was translated into the 5 selected languages. 
The translation was processed using Google Translate.
So, each non-translated English sentence was paired with the translation of its original counterpart. 
Additionally, to exclude the risk of the effect of bad translations on the performance of the model, all the source-translated sentence pairs were translation-risk predicted using ModelFront \footnote{\url{https://modelfront.com}}. 
Translations with an estimated risk score above 50\% were discarded.
In the resulting sets, the average risk varied from 24\% to 35\% among the languages.

This approach allowed us to generate more challenging negative examples for the evaluation of the model.
Unlike SemEval-17 Task 1 \cite{cer2017semeval}, where manual work of experts was involved, we were able to automatically create examples for multiple languages, including a under-resourced language such as Armenian, and then compare the detection quality of the model across the language.

\begin{table}[]
\centering
\resizebox{0.7\columnwidth}{!}{
\begin{tabular}{lccccc}
\hline
\textbf{Dataset} & Lang.  & \textbf{Pairs} & \textbf{Recall} & \textbf{Precision} & \textbf{F1} \\
\hline
$\nu$MRPC& EN-HY & 901  & 0,89   & 0,74      & 0,81\\
         & EN-RU & 1250 & 0,91   & 0,72      & 0,81\\
         & EN-ES & 1457 & 0,94   & 0,72      & 0,82\\
         & EN-FR & 1397 & 0,91   & 0,72      & 0,81\\
         & EN-DE & 1154 & 0.94   & 0.70      & 0,81\\
\hline
Neg-1    & EN-RU  & 7998  & 0,89   & 0,93      & 0,91\\
Neg-4    & EN-RU  & 18613  & 0,84   & 0,88      & 0,86\\
\hline
\end{tabular}}
\caption{Detailed analysis model scores on different benchmarks.}
\label{table:da}
\end{table}

In the following experiments, to achieve higher precision, the classification threshold of the finetuned model was chosen according to the highest $F_{0.25}$ score.

\subsection{Cross-Lingual Plagiarism Detection Benchmarks}

For the English-Russian, English-French, English-Spanish language pairs we used existing benchmarks for cross-lingual plagiarism detection and compared the results with novel and state-of-the-art results on them.
We also constructed a new test dataset for all these pairs and evaluated the method on it.
Since the scope of this work did not include the study of post-processing steps aimed at increasing the granularity score \citep{potthast2010plagdet}, we compare the methods based on precision and recall measures only.
The selected datasets and evaluation results are described below:

\textbf{Crosslang} is an automatically generated English-Russian cross-lingual plagiarism detection dataset \citep{kuznetsova2021methods}. 
This dataset is based on English and Russian Wikipedia random articles. 
Here, suspicious and corresponding source documents have similar topics.
Each suspicious document contains from 20 to 80 percent of plagiarism, the source of which is split among 1 to 10 documents.

We compare the performance of our approach with \citep{kuznetsova2021methods}, the state-of-the-art on this benchmark.
The evaluation results on CrossLang are provided in Table \ref{table:crosslang}.

\begin{table}[]
\centering
\resizebox{0.7\columnwidth}{!}{\begin{tabular}{lccc}
\hline
                              & \multicolumn{3}{c}{\textbf{Crosslang}} \\
\hline
                              & \textbf{Rec.}  & \textbf{Prec.}  & \textbf{F1} \\
\hline
Proposed method                & 0,77           & \textbf{0,86}            & \textbf{0,81}        \\
\citep{kuznetsova2021methods}  & \textbf{0,79}           & 0,83            & 0,80        \\
\hline
\end{tabular}}
\caption{Comparison of the proposed method with the state-of-the-art on CrossLang dataset for English-Russian cross-lingual plagiarism detection.}
\label{table:crosslang}
\end{table}

\textbf{\citep{Ferrero:016}}. We also used the \citep{Ferrero:016} pre-gathered subsets of different parallel corpora that are often used for different cross-language NLP tasks:
\begin{itemize}
    \item \textit{JRC-Acquis}\footnote{\href{https://ec.europa.eu/jrc/en/language-technologies/jrc-acquis}{ec.europa.eu/../jrc-acquis}} - corpus contains European law texts from Acquis Communautaire on 22 languages.
    \item \textit{Amazon Product Reviews} - a corpus of reviews of different products collected by \citep{prettenhofer2010cross}. 
    \item \textit{Conference papers} \textit{(TALN)} \citep{boudin2013taln} - a corpus that contains scientific texts extracted from conference papers that were published in two languages.
\end{itemize}

We used the same methodology as \citep{ferrero2017usingword} for evaluation, and compare the performance of our approach with the results published in their paper (Table \ref{table:ferrero}).

\begin{table}[h]
\centering
\resizebox{0.7\columnwidth}{!}
{\begin{tabular}{lcccc}
\hline
\centering
                            & \textbf{JRC-Aquis} & \textbf{APR}   & \textbf{TALN}                      \\
\hline
Proposed Method                  & \textbf{71,80{\tiny $\pm0,444$}}         & \textbf{96,67{\tiny $\pm0,387$}} & \textbf{89,72{\tiny $\pm0,474$}} \\
\citep{ferrero2017usingword} & \textbf{72,70{\tiny $\pm1,446$}}         & 78,91{\tiny $\pm1,005$} & 80,89{\tiny $\pm0,944$}  \\
\hline
\end{tabular}}
\caption{Comparison of F1-scores of the proposed method and the state-of-the-art on Ferrero et al. dataset for English-French cross-lingual plagiarism detection.}
\label{table:ferrero}
\end{table}

\textbf{PAN-PC-2011}. This corpus contains both monolingual and cross-language plagiarism cases \citep{potthast_martin_2011_3250095}. 
PAN-PC-2011 contains documents with both automatically and manually generated plagiarism cases. 
For the cross-language plagiarism detection task, this corpus includes Spanish-English and German-English document pairs. 
Some of these documents were manually corrected after an automatic translation of some parts of the texts. 
To use the dataset to evaluate our algorithm we only used the documents that were created for the external plagiarism detection task and that are used in cross-language cases.

We used the evaluation script\footnote{\href{https://github.com/pan-webis-de/pan-code/tree/master/sepln09}{github.com/pan-webis-de/../sepln09}} provided by PAN 2010 conference organizers.
The results on the Spanish-English subset are shown in Table \ref{table:pan11}.
We compare our approach against the novel state-of-the-art method proposed by \citep{roostaee2020cross}.

\begin{table}[h]
\centering
\resizebox{0.7\columnwidth}{!}{\begin{tabular}{lccc}
\hline
                        & \multicolumn{2}{c}{\textbf{PAN-PC-11 ES-EN}}\\
\hline
                        & \textbf{Recall}  & \textbf{Precision} \\
\hline
Proposed method         & \textbf{0,79}           & \textbf{0,85}       \\
\citep{roostaee2020cross} & 0,75           & 0,79       \\
\hline
\end{tabular}}
\caption{Comparison of proposed method and the state-of-the-art on PAN-PC-11 dataset's English-Spanish subset.}
\label{table:pan11}
\end{table}

\textbf{Dataset for all language pairs.} To be able to evaluate and compare the efficiency of the proposed algorithm on the 5 selected languages, we created a new dataset\footnote{\url{https://drive.google.com/drive/folders/1jnAehDCQM_u1P3wKpRMozpbpiu0xbP5E?usp=sharing}} based on Wikipedia articles, using the same algorithm as \citep{kuznetsova2021methods}.
For each language the resulting dataset contains 400 suspicious texts with plagiarised passages from a reference collection of 120,000 English texts. 

Detailed statistics and evaluation results are provided in Tables \ref{table:cl-stats} and \ref{table:cl-results} respectively.

\begin{table}[h]
\centering
\resizebox{0.7\columnwidth}{!}{
\begin{tabular}{lp{0.15\columnwidth}p{0.15\columnwidth}p{0.15\columnwidth}}
\hline
                        & \multicolumn{3}{c}{\textbf{Plagiarism fraction}}\\
\hline
\textbf{Pair}   & \textbf{Hardly \space (0-0.2)}  & \textbf{Medium (0.2-0.5)} & \textbf{Much (0.5-0.8)} \\
\hline
EN-HY    & 16.5\%  & 65.25\%   & 18.25\%  \\
EN-RU    & 15.25\% & 62.25\%   & 22.5\%  \\
EN-ES    & 7.5\%   & 59.0\%    & 33.5\%  \\
EN-FR    & 12.0\% & 60.0\%    & 28.0\%  \\
EN-DE    & 19.0\%  & 63.0\%    & 18.0\%  \\
\hline
\end{tabular}}
\caption{Plagiarism fraction per document in the proposed cross-lingual plagiarism detection dataset.}
\label{table:cl-stats}
\end{table}

\begin{table}[h]
\centering
\resizebox{0.7\columnwidth}{!}{
\begin{tabular}{lccc}
\hline
\textbf{Pair}   & \textbf{Recall}  & \textbf{Precision} & \textbf{F1} \\
\hline
EN-HY    & 0,72  & 0,73   & 0,73  \\
EN-RU    & 0,81  & 0,82   & 0,81  \\
EN-ES    & 0,90  & 0,86   & 0,88  \\
EN-FR    & 0,88  & 0,81   & 0,84  \\
EN-DE    & 0,71  & 0,64   & 0,67  \\
\hline
\end{tabular}}
\caption{Evaluation results on the proposed cross-lingual plagiarism detection benchmark.}
\label{table:cl-results}
\end{table}

\section{Discussion}

The evaluation results demonstrate that the proposed simple approach of grouping multilingual synsets based on the most frequent meaning can be very effective for cross-lingual plagiarism detection.
The method based on adapted UWN and XLM-R achieved results close to the state-of-the-art of English-French, English-Russian languages on established test datasets.

The detailed analysis results on $\nu$MRPC for the Armenian language demonstrate the effectiveness of multilingual language models in this task even for under-resourced languages.
Our approach also achieved adequate results for the overall detection task, which indicates its suitability for under-resourced languages.
In that regard, the advantages of proposed method are that it does not depend on machine translation and does not require fine-grained word sense disambiguation.
Therefore, it can easily be adapted and employed for all languages present in UWN and XLM-R.

Speaking about the limitations of the work, we need to note some points.
First, the proposed method is applicable only for the languages where lemmatization and tokenization tools exist. 
Second, Universal WordNet contains limited amount of languages, which also affects the number of applicable languages. 
Then, descent machine translation tools are required, as the method uses such a tool to supplement the “Cross-Language synonyms dictionary”.
Further, taking into account the results achieved on German texts, we can assume that the method has some issues with the languages where word formation via compounding is common.
It is worth to note that the method was tested only on the languages from Indo-European family.

\section{Conclusion}

In this work we presented a method of cross-lingual plagiarism detection that relies on openly available multilingual thesauri and pre-trained language models, and therefore can easily be adapted to many languages.
The effectiveness of the model was shown both on large languages using existing benchmarks, as well as on an under-resourced language, using newly presented datasets.

\section{Acknowledgments}

This work was supported by a grant for research centers in the field of artificial intelligence, provided by the Analytical Center for the Government of the Russian Federation in accordance with the subsidy agreement (agreement identifier 000000D730321P5Q0002 ) and the agreement with the Ivannikov Institute for System Programming of the Russian Academy of Sciences dated November 2, 2021 No. 70-2021-00142.

The authors thank Denis Turdakov, Yaroslav Nedumov, and Kirill Skornyakov for their insightful feedback.


\bibliography{mybibfile}

\end{document}